\documentclass{article}
\usepackage{ijcai19}
\usepackage{times}
\usepackage{booktabs} % For formal tables
\usepackage{balance}  % to better equalize the last page
\usepackage{graphicx} % for EPS, load graphicx instead
\usepackage{url}      % llt: nicely formatted URLs
\usepackage[T1]{fontenc}
\usepackage{listings}
\usepackage{amssymb}
\usepackage{amsbsy}
\usepackage{algorithm}
\usepackage[noend]{algpseudocode}
\usepackage{color}
\usepackage{amsmath}
\usepackage[utf8]{inputenc}
\usepackage{enumitem}
\usepackage{subfig}
\usepackage{bm}
\usepackage{multirow}
\usepackage{multicol}

\newcommand\norm[1]{\left\lVert#1\right\rVert}
\newcommand{\tp}{^{\mathrm{T}}}
\DeclareMathOperator*{\argmin}{arg\,min}
\DeclareMathOperator*{\argmax}{arg\,max}
\DeclareMathOperator{\logit}{logit}
\DeclareMathOperator{\E}{E}
\DeclareMathOperator{\betapdf}{Beta}
\DeclareMathOperator{\hcauchypdf}{Half--Cauchy}
\DeclareMathOperator{\exppdf}{Exponential}
\DeclareMathOperator{\bernoullipdf}{Bernoulli}
\DeclareMathOperator{\multinomialpdf}{Multinomial}
\DeclareMathOperator{\dirichletpdf}{Dirichlet}
\DeclareMathOperator{\categoricalpdf}{Categorical}
\DeclareMathOperator{\normalpdf}{N}
\DeclareMathOperator{\uniformpdf}{Uniform}
\DeclareMathOperator{\weibullpdf}{Weibull}
\DeclareMathOperator{\invchipdf}{Inv-\chi^2}
\DeclareMathOperator{\gammapdf}{Gamma}

\title{Human-in-the-loop Active Covariance Learning for Improving Prediction in Small Data Sets}
\author{
	Homayun Afrabandpey\footnote{Contact Author}\and
	Tomi Peltola\And
	Samuel Kaski\\
	\affiliations
	Department of Computer Science, Aalto University\\
	Helsinki Institute for Information Technology (HIIT)\\
	\emails
	firstname.lastname@aalto.fi
}
\begin{document}
	\maketitle
	
	\begin{abstract}
		Learning predictive models from small high-dimensional data sets is a key problem in high-dimensional statistics. Expert knowledge elicitation can help, and a strong line of work focuses on directly eliciting informative prior distributions for parameters. This either requires considerable statistical expertise or is laborious, as the emphasis has been on accuracy and not on efficiency of the process. Another line of work queries about importance of features one at a time, assuming them to be independent and hence missing covariance information. In contrast, we propose eliciting expert knowledge about pairwise feature similarities, to borrow statistical strength in the predictions, and using sequential decision making techniques to minimize the effort of the expert. Empirical results demonstrate improvement in predictive performance on both simulated and real data, in high-dimensional linear regression tasks, where we learn the covariance structure with a Gaussian process, based on sequential elicitation.
		% So far nobody investigates the possibility of constructing informative prior distributions with full covariance matrix by directly eliciting knowledge from an expert. The problem is challenging due to the fact that it requires elicitation of extra information about the pairs of features, the number of which could be huge in the ``small $n$, large $p$'' setting. Focusing on Bayesian linear regression, we introduce a prior elicitation model for constructing an informative multivariate Gaussian prior with a full covariance matrix (with no independence assumption among variables) for a given ``small $n$, large $p$'' problem. The prior knowledge of the domain expert is considered to be on similarities of the features. Evaluations of the proposed method in experiments with simulated and real data show improved prediction accuracy with small effort from the expert.
	\end{abstract}
	
	\section{Introduction}
	
	Data sets with a small number of observations $n$ and a large number of variables $p$, a.k.a ``small $n$, large $p$'', are common in many applications and %. These data sets differ from the ``big'' data sets that drove the evolution of data analysis during the last decades and
	pose big challenges for statistical methods \cite{fan2006statistical}. % and limit their performance \cite{donoho2009observed}. 
	Potential approaches to alleviate these challenges %of ``small $n$, large $p$'' 
	are either to provide more samples, which could be very expensive or even impossible in some applications, or to regularize models using additional sources of information. An example of such an additional source is human expert who may have substantial subject-specific knowledge that can be used to improve data analysis.
	
	In Bayesian statistics, expert-provided information can be formulated as a prior distribution over the parameters to regularize the model. Extracting and incorporating experts' knowledge into statistical models is known as \textit{knowledge elicitation}. When designing an elicitation method, two important choices will affect the success. First is what is assumed of the knowledge of the expert: only domain expertise or additional statistical expertise. In Bayesian statistics, most of the existing knowledge elicitation approaches require the domain expert to be expert, or at least knowledgeable, in statistics as well. The second choice is the assumptions made on the content of knowledge that is elicited, which also translates into the structure of the model used for the elicitation. Most existing methods for eliciting importance of variables for prediction tasks assume independence among the parameters to make the elicitation easier, at the cost of ignoring expert knowledge on dependencies.
	
	In this paper, we propose a method for elicitation that avoids both problems: only domain expertise is required, and the underlying assumptions are generalized from independence of the parameters to pairwise dependencies. Technically, we formulate the problem as a sequential decision-making task of choosing which queries to ask to maximally improve predictions with minimal expert effort. The queries will be about pairwise relationships of features, to expand beyond the earlier assumption of their independence. 
	
	With experiments on synthetic and real data, we demonstrate that the approach is able to improve the predictive performance by constructing a more informative covariance matrix using expert feedback. 
	
	\textbf{Contribution.} In summary, the main contributions of the paper are: \textbf{(1)} A method for learning a prior with a full covariance matrix (without an independence assumption among features) using information provided by a domain expert, \textbf{(2)} an algorithm for actively eliciting expert's knowledge on pairwise similarities of the features, to maximally improve predictive performance while minimizing the expert's workload.
	
	\section{Related Work}\label{Sec:2}
	
	There exists a rich literature on knowledge elicitation for improving the performance of statistical models. %In the following, we give a brief overview of the papers that are the most relevant to our work.
	A number of studies have proposed to elicit scores or point estimates of an unknown parameter or quantity of interest directly from the expert \cite{garthwaite2008use,haakma2011pmd4,o1998eliciting}. %In \cite{kadane1980interactive} the authors proposed to elicit from experts their subjective probability distributions of predicted responses in a linear regression model. Similar approaches have been proposed for logistic regression \cite{al2006quantifying} and other generalized linear models \cite{garthwaite2013prior,hosack2017prior}. 
	These approaches typically either assume that the expert has a high level of statistical training, to be able to assess and formulate prior distributions, or is assisted by a human facilitator in an elicitation session. The main goal is accuracy of the knowledge instead of time-efficiency, which makes the elicitation burdensome. 
	%They could be used to complement our approach. Some studies proposed to elicit scores or point estimates (such as the mean and/or mode) of an unknown parameter directly from the expert. Examples include \cite{o1998eliciting, haakma2011pmd4, garthwaite2008use} where the expert is asked to specify the mode (the most likely values) and/or median of a quantity of interest. The problem with these studies is that such extreme knowledge is rarely available to the expert. Alternative to these approaches is to develop distribution (instead of point estimates) for the unknown parameter of interest. In \cite{kadane1980interactive} the authors proposed to elicit from experts their subjective probability distributions of predicted responses in a linear regression model. Similar approaches proposed for logistic regression \cite{al2006quantifying} and other generalized linear models \cite{garthwaite2013prior,hosack2017prior}. These methods still suffer from unavailability of such knowledge to domain experts. We argue that elicitation methods must permit non-statisticians to contribute their judgments since in most applications, experts are not knowledgeable in statistics. Furthermore, they must mitigate the required elicitation workload to a manageable level.
	In an attempt to make knowledge elicitation more automatic and resource-efficient, Micallef et al.~\shortcite{Micallef2017Interactive} proposed an interactive approach to elicit the expert's knowledge on the relevance of individual features to improve the predictive accuracy of an underlying model. They used a sequential decision making method, multi-armed bandits, to guide the interaction toward features that would likely benefit from the expert's input. Daee et al.~\shortcite{daee2017knowledge} formulated knowledge elicitation as a probabilistic inference problem where expert knowledge about individual features is sequentially extracted to improve predictions in a sparse linear regression task. The work was extended by Sundin et al.~\shortcite{sundin2018improving} by considering information about the direction of a relevant feature. In all these works, independence among features is assumed in the prior distribution. These approaches could be used to complement our method. %In an attempt to make prior elicitation easier and more feasible for non-statisticians, Micallef et al. \cite{Micallef2017Interactive} proposed an interactive approach to elicit expert's prior knowledge on the relevance of individual features and use it to improve the prediction accuracy of an underlying model. This approach assumes an expert model which queries only the features with highest Upper Confidence Bound (UCB) to guide the interaction toward features that would likely benefit from expert's input. The prediction improvement is obtained by modifying the variance of the prior distribution using expert's feedback. In another work, Daee et al. \cite{daee2017knowledge} formulated knowledge elicitation as a probabilistic inference problem where expert knowledge is sequentially extracted to improve predictions in a sparse linear regression task. The authors assumed two types of prior knowledge for the expert; either the expert has knowledge on exact value of the regression coefficient of a variable or she has knowledge on the relevance of a feature to the prediction task. 
	
	In a similar scenario, Xiao et al.~\shortcite{xiao2018optimal} proposed a knowledge elicitation approach for extracting expert knowledge on influential relationships between pairs of variables in a directed acyclic graph, i.e., whether a variable $a$ is likely to be up- or down-stream of variable $b$. Their goal, however, is not prediction but to learn the structure of the graph. Afrabandpey et al.~\shortcite{afrabandpey2017interactive} proposed an interactive method for eliciting knowledge on pairwise feature similarities based on an interactive display and used it to improve predictive performance of linear regression. This work is similar to ours, but differs in two important aspects: First, their method is passive - the queries do not change according to what has been elicited so far - which makes the feedback giving process exhaustive for the expert. Second, the expert feedbacks are only post-hoc connected to the prediction problem through a metric learning process, %(a two-dimensional projection of the features),
	while our approach adopts expert feedback directly by constructing an informative prior distribution for the underlying model.
	
	Our method builds upon two further works. Krupka and Tishby~\shortcite{krupka2007incorporating} studied cases where there are meta-features that capture prior knowledge on features. Using these meta-features, they proposed a method to improve generalization of a classification model by constructing a prior covariance (or a Gaussian process) on the parameters without independence assumptions among the variables. Their approach, however, is not a human-in-the-loop system and does not elicit expert knowledge. Yang et al.~\shortcite{yang2007bayesian} proposed a Bayesian framework for actively learning a similarity metric among samples using pairwise constraints labeled by a human agent. The method aims to improve pairwise distances by querying the unlabeled pairs of samples with the greatest uncertainty in relative distance. In contrast, our approach is designed to optimize the predictive performance of the model using feedback on pairs of features and we query pairs based on their expected improvement on predictive performance. %In a different scenario, \cite{krupka2007incorporating} studied cases where there are meta-features that capture some prior knowledge on features. Using these meta-features, they proposed a method to improve generalization of a classification model by constructing a prior covariance on the parameters without independence assumption among the variables. The approach, however, is not developed as a human-in-the-loop approach and it is not clear how a human expert can contribute in the construction of the model. Related to our work is the active distance learning approach by Yang et. al. \cite{yang2007bayesian}, where they proposed a Bayesian framework for learning a similarity metric among samples using pairwise constraints labeled by a human agent. The method aims to improve pairwise distances by querying unlabeled pairs of samples with greatest uncertainty in relative distance. In contrast, our approach is designed to optimize the predictive performance of the model using feedback on pairs of features and we query pairs based on their expected improvement on predictive performance.
	
	\section{Proposed Approach}\label{Sec:4}
	
	We consider the task of learning a probabilistic predictive function $y = h_{\bm{\theta}}(\bm{x})$, parametrized by $\bm{\theta}$, given a data set $\mathcal{D} = \{(y_i, \bm{x}_i); i = 1,\ldots, n\}$ of $n$ outcomes $y_i$ and feature vectors $\bm{x}_i = [x_{i1}, \ldots, x_{ip}]\tp$ pairs. We assume that $n$ is small and cannot be increased, so that it is worthwhile to spend effort on collecting and encoding prior knowledge or other external information. Here, we particularly focus on using knowledge elicitation from domain expert to learn the dependencies between the elements of $\bm{\theta} = [\theta_1, \ldots, \theta_L]\tp$, %to allow the learning of $\bm{\theta}$ from $\mathcal{D}$ to share statistical strength between the elements 
	where $L$ is the total number of parameters of the predictive function. This approach is complementary to many of the previous prior and expert knowledge elicitation approaches.%In Section~\ref{Sec:5}, we will demonstrate the approach with experiments in the case of linear regression, $f_{\bm{\beta}}(\bm{x}) = \bm{\beta}\tp \bm{x}$, with $\bm{\theta} = \bm{\beta}$ being the regression coefficients, in settings where $n$ is small compared to $p$.
	
\begin{figure}[bt!]
		\centering
		\includegraphics[width = \columnwidth]{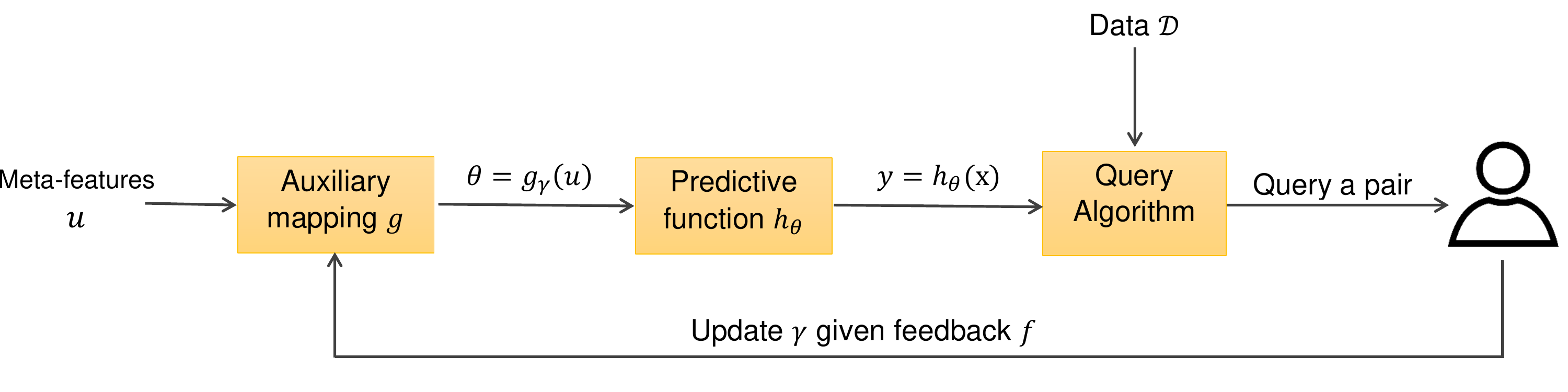}
		\caption{Components of the proposed approach. To improve the predictive function $h_{\bm{\theta}}$, the auxiliary mapping $g_{\bm{\gamma}}$ models dependencies between the parameters $\bm{\theta}$ using the meta-features $\bm{u}$ and the current value of $\bm{\gamma}$. The query algorithm optimizes the workload of the expert by choosing queries that maximize the expected gain in the predictive performance. The expert feedback $f$ then updates $\bm{\gamma}$.}
		\label{model_comp}
	\end{figure}

	In addition to the data $\mathcal{D}$, we assume the availability of meta-features $\bm{u}_l \in \mathbb{R}^d$ for each parameter $\theta_l$. Our approach to improve the predictive model $h_{\bm{\theta}}$ is then formulated in three components as shown in Figure \ref{model_comp}: (1) An auxiliary mapping $g$ from the meta-features $\bm{u}_l$ to the parameters $\theta_l$, $\theta_l = g_{\bm{\gamma}}(\bm{u}_l)$, with parameters $\bm{\gamma}$. (2) An observation model for expert feedback that encodes the expert knowledge to information about $g$ and consequently to the dependencies between the elements of $\bm{\theta}$. (3) A query algorithm that chooses questions to the expert to optimize the effort of the expert for improving the predictions. Each component is described in more detail in the following sections.
	% \begin{figure}[bt!]
	%     \centering
	%     \includegraphics[width = \columnwidth]{model_comp.pdf}
	%     \caption{Components of the proposed approach. Given the meta-features and the current value of $\gamma$, the auxiliary mapping $g$ constructs the input to the query algorithm which then chooses a pair to be queried to the expert. The expert feedback then updates $\gamma$ and the procedure iterates until some stopping criteria. The focus of the figure is on elicitation part of the approach.}
	%     \label{model_comp}
	% \end{figure}
	
	To concretize the approach, consider as an example a sentiment analysis task where the goal is to predict a rating assigned by a reviewer to a product from the text of the review. The training data is the set of all reviews given by different reviewers together with their corresponding ratings. The feature vector $\bm{x}$ describes occurrence of keywords in the review text (bag-of-words). Assume linear regression for prediction, $h_{\bm{\beta}}(\bm{x}) = \bm{\beta}\tp \bm{x}$, with $\bm{\theta} = \bm{\beta}$ being the regression coefficients. A simple approach to constructing a meta-feature vector $\bm{u}_l$ for each keyword $l$ is to take the vector of occurrences of the keyword in the training set reviews. This implies that features that do not occur in the same documents are dissimilar and vice versa. Using a Gaussian process as the auxiliary mapping $g$ then corresponds to learning a full covariance matrix for the prior of the regression coefficients $\bm{\beta}$ based on the meta-features. Knowledge about the pairwise similarity of the role of the keywords in the rating prediction is sequentially queried from the expert. The rationale is that the expert has knowledge about the correlation among regression coefficients of at least some keywords, i.e., either that two keywords positively affect the output (both cause an increase in the rating) or they negatively affect it (decrease the rating). The extracted knowledge will then be incorporated into the model to improve the prior covariance matrix and thus predictions. The case of prediction task with linear regression is described in detail in Section \ref{Sec:4.1}.
	
	\subsection{Modelling Parameter Dependencies}
	
	The auxiliary mapping $g$ is used to model the dependencies between the parameters $\bm{\theta}$. We model it with a Gaussian process with a zero mean and covariance function $k_{\bm{\gamma}}$, where $\bm{\gamma}$ is a vector of parameters \cite{krupka2007incorporating,rasmussen2006gaussian}. In particular, this implies that $\bm{\theta} \sim \mathcal{N}(\bm{0}, \bm{K})$, where $\bm{K}$ is a covariance matrix (kernel). The element $K_{ij} = k_{\bm{\gamma}}(\bm{u}_i, \bm{u}_j)$ describes the prior covariance between the parameters $\theta_i$ and $\theta_j$ based on the similarity of their meta-features $\bm{u}_i$ and $\bm{u}_j$ (according to the covariance function).
	
	To illustrate the effect of the auxiliary mapping on learning $\bm{\theta}$, consider a two-dimensional special case, with $\theta_1$ and $\theta_2$. %, and first its extremes.
	If the prior correlation implied by $\bm{K}$ is equal to one, then $\theta_1 = \theta_2$; that is, the $g$ maps the two meta-feature vectors $\bm{u}_1$ and $\bm{u}_2$ to the same point and, effectively, there is only one parameter to be learned from the data set $\mathcal{D}$. If the prior correlation is zero, learning information about one parameter has no effect on the other, and we need to estimate both of the parameters from $\mathcal{D}$. Of course, often we would be somewhere between these two extremes, such that some statistical strength can be shared between learning of the two parameters.
	
	The covariance function $k_{\bm{\gamma}}$ defines many of the properties (smoothness, flexibility, magnitude, etc.) of the auxiliary mapping based on the meta-features $\bm{u}_i$ and $\bm{u}_j$. Many different covariance functions are possible (see, e.g., \cite{rasmussen2006gaussian}). We use the Gaussian kernel
	\vspace*{-0.1cm}
	\begin{equation}\label{eqn:gaussiankernel}
		%\bm{K} = \left[ k_{ij} = \exp \left( - \frac{\norm{\bm{u}_{i} - \bm{u}_{j}}_{\bm{A}}^{2}}{2} \right) \right]_{i,j = 1}^{L},
		k_{\bm{\gamma}}(\bm{u}_{i}, \bm{u}_{j}) = \exp \left( - \frac{\norm{\bm{u}_{i} - \bm{u}_{j}}_{\bm{A}_{\bm{\gamma}}}^{2}}{2} \right),
	\end{equation}
	where $\bm{A}_{\bm{\gamma}} \in \mathbb{R}^{d \times d}$ is a diagonal matrix defining the distance metric with elements $\gamma_l$, such that $\norm{\bm{u}_{i} - \bm{u}_{j}}_{\bm{A}_{\bm{\gamma}}}^{2} = \sum_l \gamma_l (u_{il} - u_{jl})^2$. Here, the $\bm{\gamma}$ are inverse length scales: they determine how influential each dimension of the meta-feature is in the distance metric; $\gamma_i = 0$ means that the $i^{th}$ dimension of the meta-features does not influence the distance (all features are equal w.r.t.\ this descriptor). Increasing the value from $0$ means that it is increasingly important in defining the distance. %: large value of $\gamma_i$ means that even small distance between two features in this dimension of the meta-feature makes the features dissimilar. 
	Inference about $\bm{\gamma}$ is done by defining a half-normal prior distribution $\bm{\gamma} \sim \mathcal{N}^{+}\left( \gamma_0\bm{1}_{d \times 1}, \delta \bm{I}_{d \times d} \right)$. The $\gamma_0$ and $\delta$ are constant values.% defined before running the experiment.
	
	The auxiliary mapping $g$ defines a prior on the parameters $\bm{\theta}$, based on the meta features. It would be attractive to learn $g$ from the data $\mathcal{D}$, but that would require more data than available in the ”small $n$, large $p$” case we are targeting. That is why we turn next to learning it based on knowledge elicitation instead.
	%While the auxiliary mapping $g$ defines a prior distribution on the parameters $\bm{\theta}$, based on the meta-features and the chosen covariance function, the data set $\mathcal{D}$ may contain limited information to learn the mapping and its parameters $\bm{\gamma}$, since $\bm{\theta}$ would often not be directly observable and thus some levels of the model's hierarchy away from the observed data. As a solution, % to this,
	%we next describe how expert knowledge on the pairwise similarity of the elements of $\bm{\theta}$ can be incorporated into the model.
	
	\subsection{Feedback Model}
	
	A feedback model is used to incorporate expert's knowledge into the model. In this work, we consider feedback on the pairwise similarities of the roles of the parameters $\bm{\theta}$ in the prediction. For example, if the $\bm{\theta}$ are linear regression weights, the expert tells the model which pairs of features affect the prediction output similarly (positively and/or negatively). Expert's feedback will be either ``Similar'', ``Dissimilar'', or ``I don't know''. The latter will be ignored by the model and has no effect on prior covariance and consequently on the predictions. Similar to \cite{yang2007bayesian}, we define the following likelihood for the feedbacks:
	\begin{equation}
		f_{ij} \sim \mbox{Bernoulli}(q),
	\end{equation}
	where $f_{ij}$ is the feedback given on the similarity of the features $i$ and $j$ ($f_{ij} = 1$ for similar and $f_{ij} = -1$ dissimilar feedback), $q = \frac{1}{1 + \mbox{exp}\left( \bm{\gamma}^{T}\bm{w}_{ij} - \xi \right)}$ is the parameter of the Bernoulli distribution,
	$\bm{w}_{ij} = \left[ \bm{u}_{i} - \bm{u}_{j} \right]^{2}$, and $\xi$ is the threshold determining the similarity/dissimilarity of the two features. Since feedbacks are binary, the feedback model is basically defined as a logistic regression of the distances among features $\bm{w}_{ij}$. Two features are more likely to be similar, $f_{ij} = 1$, only when their distance is less than the threshold $\xi$ and vice versa. To avoid setting the value of the threshold manually, we define a half-normal prior distribution over it, $\xi \sim \mathcal{N}^{+}\left( m_{\xi}, \sigma^{2}_{\xi} \right)$.
	
	\subsection{Query Algorithm}\label{Sec:4.4}
	
	Given the set of pairs of features, we would like to query first the similarity of the pair that will give us maximal improvement on predictive performance. Our query algorithm chooses the pair of features that maximizes the expected utility, which is a standard criterion in decision theory. The utility is task dependent - in our prediction task it is natural to choose the utility to be information gain of the predictions. The same rationale was earlier used successfully by \cite{daee2017knowledge}, for querying individual variables. The larger the information gain, the greater impact the feedback will have on the posterior predictive distribution. This is defined as the expected Kullback--Leibler divergence (KL) between the posterior predictive distribution before seeing the feedback and after giving the feedback:
	\begin{equation}
		\centering
		\mathbb{E}_{\pi(\Tilde{f}_{ij} \mid \mathcal{D},\mathcal{F})}\!\!\!\left[\!\sum_{k}\! \mbox{KL}\!\!\left[ \pi\!\left(\Tilde{y}\!\mid\! \bm{x}_{k},\mathcal{D},\mathcal{F},\Tilde{f}_{ij}\!\right)\!\!\parallel\!\! \pi\!\left(\Tilde{y}\!\mid\! \bm{x}_{k},\mathcal{D},\mathcal{F}\right)\!\right]\!\! \right]
		\label{Eq9}
	\end{equation}
	%\begin{multline}\label{Eq9}
	%    \arg\max_{(ij) \notin F}\mathbb{E}_{p(\Tilde{f}_{ij} \mid D,F)}\left[ \sum_{k} \mbox{KL}\left[ p\left( \Tilde{y} \mid \bm{x}_{k},D,F,\Tilde{f}_{ij} \right) \parallel p\left( \Tilde{y} \mid \bm{x}_{k},D,F \right) \right] \right]
	%\end{multline}
	%, i.e $p\left(\Tilde{y} \mid \bm{\Tilde{x}}, D, F \right)$,
	%, i.e. $p\left(\Tilde{y} \mid \bm{\Tilde{x}}, D, F, \Tilde{f_{ij}} \right)$
	where $\mathcal{F}$ is the set of feedbacks given up to the current step, and the summation goes over the training data. The expectation is over the unknown feedbacks, given everything observed so far. The posterior predictive distribution of the feedback,  $\pi\!\left(\Tilde{f_{ij}}\mid \mathcal{D},\mathcal{F}\right)$, is obtained using the data and all the feedback given by the expert up to the current iteration.
	
	\subsubsection{Computational Complexity}\label{Sec:4.5}
	
	In each iteration, we need to compute the utilities for all pairs of features, except those to which the expert already gave feedback in the previous iterations. The complexity is $O(p^2)$ which is expensive if $p$ is large. To reduce this complexity, we used the idea presented in \cite{xiong2015active}: we construct a smaller pool of the pairs by randomly sampling from the original set of feature pairs, excluding the pairs to which the expert gave feedback previously. We then select the most informative pair from among the pairs in the smaller pool based on their utilities. Denoting the randomly selected pool by $R_{p}$, the objective function for the query algorithm will change to
	\begin{equation}\label{Eq10}
		(ij)^{*} = \arg\max_{(ij) \in R_{p}} (\mbox{Expected Information Gain}),
	\end{equation}
	where the expected information gain is computed using Equation~\ref{Eq9}. Although the optimal pair of $R_{p}$ might not necessarily be optimal for the full set, the amount of degradation in the performance is negligible. This is due to the fact that in real data sets with large number of features, there will be significant redundancies among the pairs. The complexity reduces from $O(p^{2})$ to $O(\mid\! R_{p}\!\mid)$, where $\mid\!R_{p}\!\mid$ denotes the cardinality of $R_{p}$.
	
	The above strategy is effective only if the selected pair from $R_p$ can be guaranteed to be near optimal with high probability. A  pair is near-optimal if it is among the $\epsilon$  top-ranked pairs in the original set, % of pairs according to the measure showed in Equation~\ref{Eq9}, 
	where $\epsilon$ is a small scalar (e.g. 0.001). The near-optimality of the selected pair from the smaller pool can be proven by the following proposition:
	
	\textbf{Proposition 1.} The best pair in $R_{p}$ according to Equation~\ref{Eq10} is among the $\epsilon$ top-ranked pairs of the original set with probability $1-(1-\epsilon)^{\mid\!R_{p}\!\mid}$.
	
	\textit{Proof.} Based on the definition of near-optimal pairs, the probability that a pair does not belong to the near-optimal pairs is $1-\epsilon$. Since the size of the smaller pool is $\mid\!R_{p}\!\mid$, the probability of obtaining a near optimal query is $1-(1-\epsilon)^{\mid\!R_{p}\!\mid}$.
	
	By setting $\mid\!R_{p}\!\mid$ to a reasonable multiple of $p$, e.g. $10p$, we reduce the complexity from $O(p^2)$ to $O(p)$, while with high probability (probability of $\sim 0.99$ for a moderate size $p = 400$), the selected pair from $R_{p}$ will be among the top $0.1$ percent pairs in the original set. To increase computational speed, we implemented computation of the utilities in parallel. This is straightforward since utilities of different pairs are independent.
	
	\subsection{Application to Linear Regression}\label{Sec:4.1}
	
	The described approach is general and could, for example, be applied directly to Gaussian process regression where $\bm{\theta}$ would be the predictive function evaluations at the training data points.
	%, or to model joint sparsity patterns in sparse linear regression similar to group-wise regularization methods, where $\bm{\theta}$ would correspond to the local sparsity parameters of the regression weights.
	In this paper, we apply the method to the important case of linear regression with ``small $n$, large $p$'' data sets, which occurs in a wide array of practical applications. In particular, we take $\bm{\theta} = \bm{\beta}$ to be the regression weights. We assume availability of an expert who has prior knowledge on the pairwise similarities of the role of the features in the prediction task.
	
	% Given a matrix of predictors $\bm{X} \in \mathbb{R}^{n \times p}$ and a vector of dependent variables, $\bm{y} \in \mathbb{R}^{n \times 1}$ where $n \ll p$, the task is to learn a linear model for the relationship between $\bm{X}$ and $\bm{y}$. The model will then be used to predict the corresponding output for a new input $\Tilde{\bm{x}} \in \mathbb{R}^{p\times 1}$. The model is defined as
	% \begin{equation}
	%     \bm{y} = \bm{X \beta} + \bm{\epsilon},
	% \end{equation}
	% where $\epsilon \sim \mathcal{N}(\bm{0}, \sigma^{2}\bm{I})$ is the vector of residuals, $\bm{I} \in \mathbb{R}^{n \times n}$ is the identity matrix, and $\bm{\beta} \in \mathbb{R}^{p \times 1}$ is the regression coefficient vector to be inferred. With these assumptions, $\bm{y \mid X, \beta},\bm{\sigma}^2 \sim \mathcal{N}\left( \bm{X\beta}, \sigma^{2}\bm{I} \right)$. The goal is to learn the posterior distribution of $\bm{\beta}$ given the training data and the expert feedback.
	The linear regression model is defined as
	\begin{equation}
		\bm{y} = \bm{X \beta} + \bm{\epsilon},
	\end{equation}
	where $\bm{X} \in \mathbb{R}^{n \times p}$ is the matrix of predictors, $\bm{y} \in \mathbb{R}^{n \times 1}$ is the vector of all dependent variables, $\epsilon \sim \mathcal{N}(\bm{0}, \sigma^{2}\bm{I})$ is the vector of residuals, %$\bm{I} \in \mathbb{R}^{n \times n}$ is the identity matrix, 
	and $\bm{\beta} \in \mathbb{R}^{p \times 1}$ is the regression coefficient vector to be inferred. With these assumptions, $\bm{y \mid X, \beta},\bm{\sigma}^2 \sim \mathcal{N}\left( \bm{X\beta}, \sigma^{2}\bm{I} \right)$. The goal is to learn the posterior distribution of $\bm{\beta}$ given the training data and the expert feedback. The proposed approach assumes the following prior distribution for $\bm{\beta}$:
	\begin{equation}
		\bm{\beta} \mid \sigma^2, \tau^2, \bm{\gamma} \sim \mathcal{N}\left( \bm{0}, \sigma^{2} \tau^{2} \bm{K} \right),
	\end{equation}
	where $\bm{K}$ is the covariance matrix defined by the Gaussian covariance function $k_{\bm{\gamma}}$ and we have also introduced a scalar magnitude parameter $\tau^2$. The expert knowledge affects the predictions through its effect on this covariance matrix.
	
	To complete the model formulation, we define inverse gamma distributions for the hyper-parameters $\sigma^{-2} \sim \mbox{Gamma} \left( a_{\sigma}, b_{\sigma} \right)$ and $\tau^{-2}\!\sim\!\mbox{Gamma} \left( a_{\tau}, b_{\tau} \right)$, where $a_{\sigma}$, $b_{\sigma}$, $a_{\tau}$ and $b_{\tau}$ are constant values. 
	Setting the values of the hyper-parameters is discussed in Section \ref{Sec:5}. Plate diagram of the model is provided in appendix C.
	
	\subsection{Posterior Inference and Prediction}\label{Sec:4.3}
	
	Parameters of the proposed model are $\{ \bm{\beta}, \sigma^{2}, \tau^{2}, \bm{\gamma}, \xi \}$. Due to the complexity of the model, there is no closed-form posterior distribution. Instead, we obtain maximum a posteriori (MAP) estimates of the parameters by implementing the model in the probabilistic programming language Stan \cite{carpenter2016stan} (version 2.17.3, codes in the supplementary material). 
	To reduce computational burden of computing the MAP estimate of $\bm{\beta}$, which requires the inversion of the covariance matrix $\bm{K}$, %at each iteration when the feedback is given by the expert. Since we assume that our data set is ``small $n$, large $p$'', inverting the matrix is computationally demanding and makes iterations very slow.
	we marginalized the likelihood over $\bm{\beta}$ and $\sigma^{2}$. This results in a Student's t-distribution with $2a_{\sigma}$ degrees of freedom for the marginal likelihood, $\bm{y} \sim \mbox{MVSt}_{2a_{\sigma}}\left( 0, \frac{b_{\sigma}}{a_{\sigma}}\bm{\Sigma} \right)$, with $\bm{\Sigma} = \bm{I} + \tau^{2}\bm{X} \bm{K}\bm{X}^{T}$. 
	% \begin{equation}
	%     \bm{y} \sim \mbox{MVSt}_{2a_{\sigma}}\left( 0, \frac{b_{\sigma}}{a_{\sigma}}\bm{\Sigma} \right),
	% \end{equation}
	The matrix $\bm{\Sigma}$ is of dimension $n\times n$ which is much smaller than $p \times p$. More details on the marginal likelihood derivation are provided in appendix B. %The Stan model provides us point estimates of $\bm{K}$ and $\tau^{2}$ (alternatively we can learn $\tau^{2}$ using cross-validation), where $\bm{K}$ obtained using a MAP estimate of $\bm{\gamma}$ and Equation~\ref{eqn:gaussiankernel}. 
	MAP estimate of $\bm{\gamma}$ is used to compute $\bm{K}$. The joint posterior distribution of $\bm{\beta}$ and $\sigma^{2}$ is then obtained as a Normal-Inverse-Gamma distribution, $\mbox{NIG}(\bm{\mu}^{*},\bm{\Sigma}^{*},a^{*},b^{*})$, with parameters
	\vspace*{-0.2cm}
	\begin{align*}
		\bm{\mu}^{*} &= \left( \left( \tau^{2}\bm{K} \right)^{-1} + \bm{X}\tp\bm{X} \right)^{-1}\bm{X}^{T}\bm{y}, \\
		\bm{\Sigma}^{*} &= \left( \left( \tau^{2}\bm{K} \right)^{-1} + \bm{X}\tp \bm{X} \right)^{-1}, \\
		a^{*} &= a_{\sigma} + \frac{N}{2}, \\
		b^{*} &= b_{\sigma} + \frac{1}{2}\left[ \bm{y}\tp\bm{y} - (\bm{\mu}^{*})\tp(\bm{\Sigma}^{*})^{-1}\bm{\mu}^{*} \right].
	\end{align*}
	% The marginal posterior distributions of each parameter $\bm{\beta}$ and $\sigma^2$ can be computed easily as
	% \begin{align}
	% \begin{split}
	%     \pi(\bm{\beta} \mid \bm{y}) &= MVSt_{2a_{\sigma}^{*}}\left(\bm{\mu}^{*}, \bm{\Sigma}^{*} \right),
	% \end{split} \\
	% \begin{split}
	%     \pi(\sigma^{2} \mid \bm{y}) &= \mbox{IG}(a_{\sigma}^{*}, b_{\sigma}^{*}).
	% \end{split}
	% \end{align}
	Finally, predictions for the future input $\Tilde{\bm{x}} $ can be done using the posterior predictive distribution:
	\begin{align}
		\pi(\Tilde{y} \mid \bm{y}) &= \int \pi(\Tilde{y} \mid \bm{\beta}, \sigma^{2})\pi(\bm{\beta}, \sigma^{2} \mid \bm{y}) d\bm{\beta}d\sigma^{2} \nonumber \\
		&= \mbox{t}_{2a^{*}} \left( \Tilde{\bm{x}}^{T} \bm{\mu}^{*}, \frac{b^{*}}{a^{*}} (1 + \Tilde{\bm{x}}^{T} \bm{\Sigma}^{*} \Tilde{\bm{x}}) \right),
	\end{align}
	where $t_{2a^{*}}$ denotes the univariate Student's t-distribution with $2a^{*}$ degrees of freedom. Details on computing the KL divergence of Equation \ref{Eq9} for the linear regression model are provided in appendix A.
	\iffalse
	\begin{multline}
		p(\Tilde{y} \mid y) = \int p(\Tilde{y} \mid \bm{\beta}, \sigma^{2})p(\bm{\beta},\sigma^{2}\mid y)d\bm{\beta}d\sigma^{2} = \\
		\mbox{MVSt}_{2a^{*}}\left( \Tilde{\bm{x}}^{T}\bm{\mu}^{*}, \frac{b^{*}}{a^{*}}\left( 1 + \Tilde{\bm{x}}^{T}\bm{\Sigma}^{*}\Tilde{\bm{x}} \right) \right)
	\end{multline}
	\fi
	
	\section{Experimental Results}\label{Sec:5}
	
	In this section, the performance of the proposed method is evaluated in several ``small $n$, large $p$'' regression problems on both simulated and real data with a simulated user. As far as we know, no other method has been proposed for precisely the same task, i.e., improving generalization by constructing a prior distribution using feedback on pairs of features. Therefore, we compare with two natural alternatives:
	\begin{itemize}[noitemsep,topsep=0.5ex]
		\item[-] random query,
		\item[-] non-sequential version of our algorithm, which computes the utilities once, before observing user feedback, and never updates the utilities.
	\end{itemize}
	\begin{figure*}[ht!]
		\centering
		\subfloat[][]{
			\includegraphics[scale=.3]{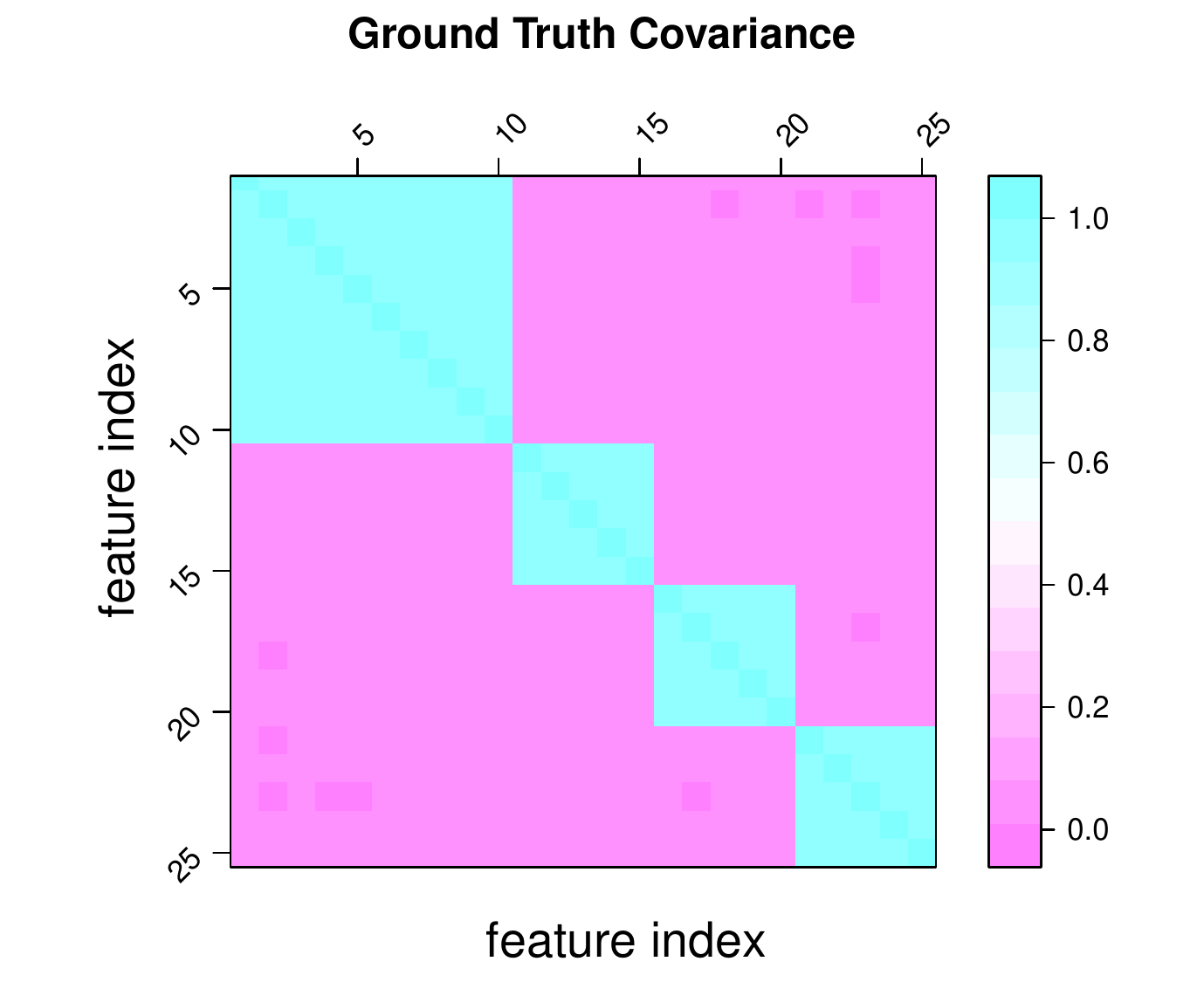}
		}\hspace{-0.2em}
		\subfloat[][]{
			\includegraphics[scale=.3]{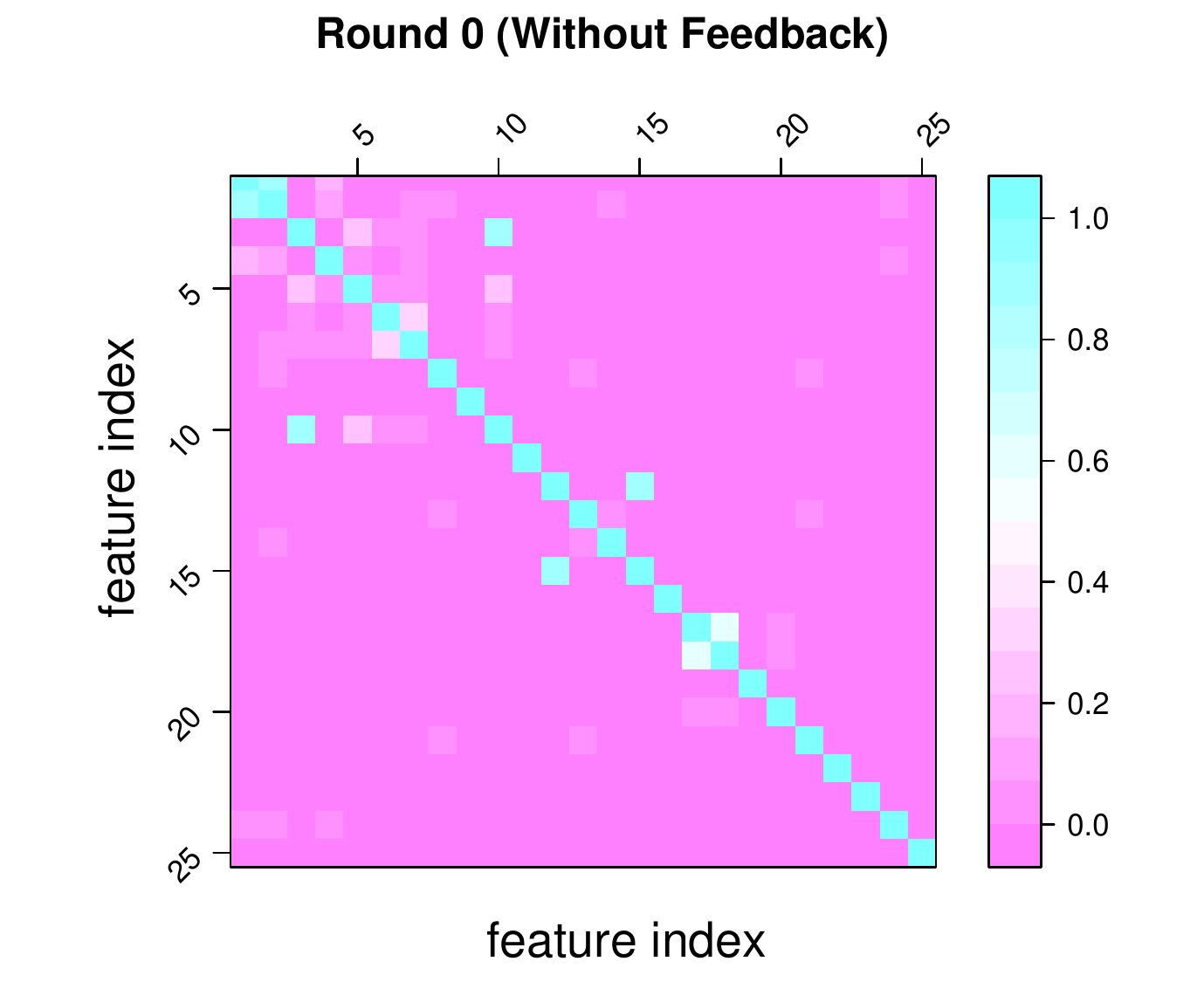}
		}\hspace{-0.2em}
		\subfloat[][]{
			\includegraphics[scale=.3]{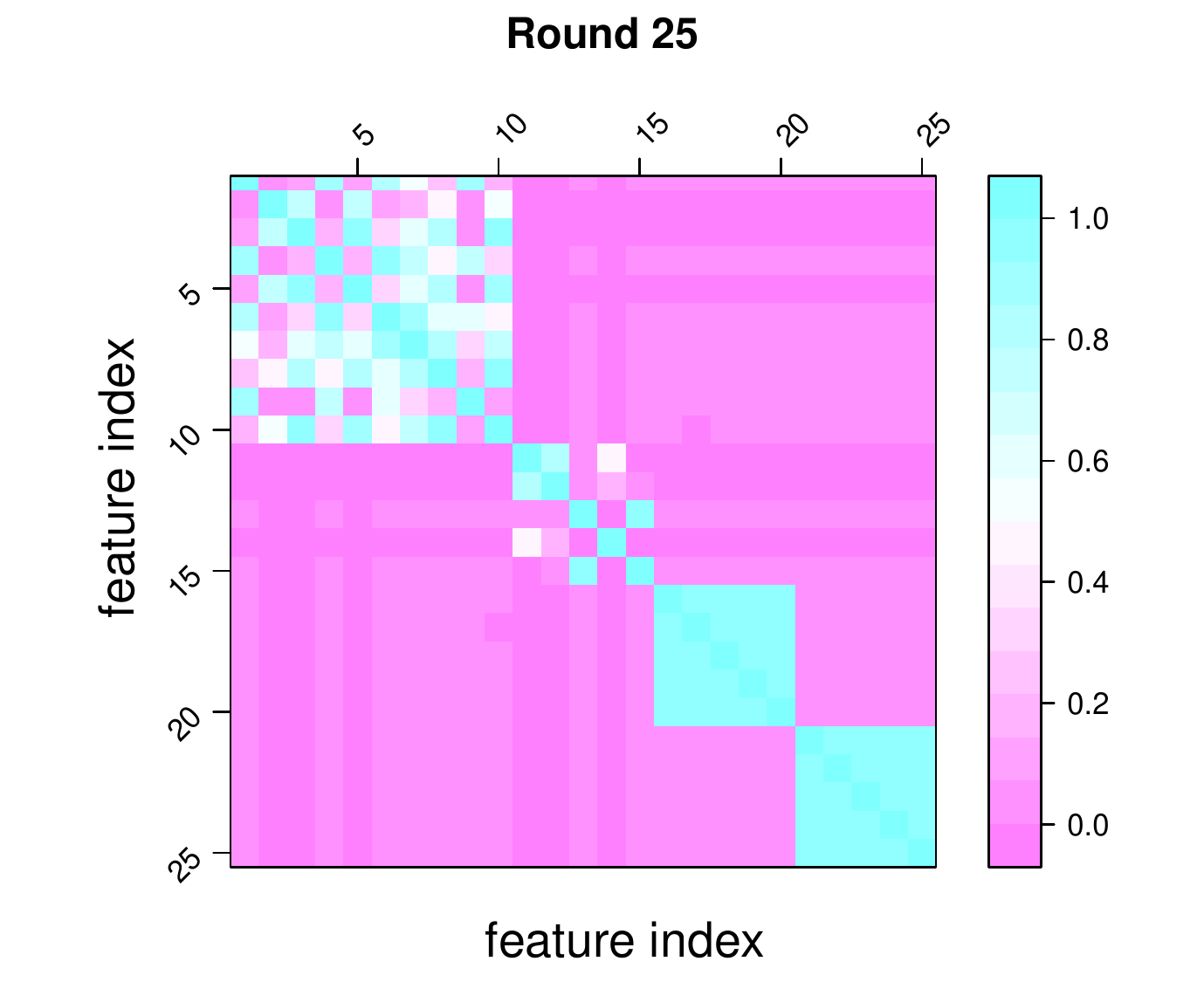}
		}\hspace{-0.2em}
		\subfloat[][]{
			\includegraphics[scale=.3]{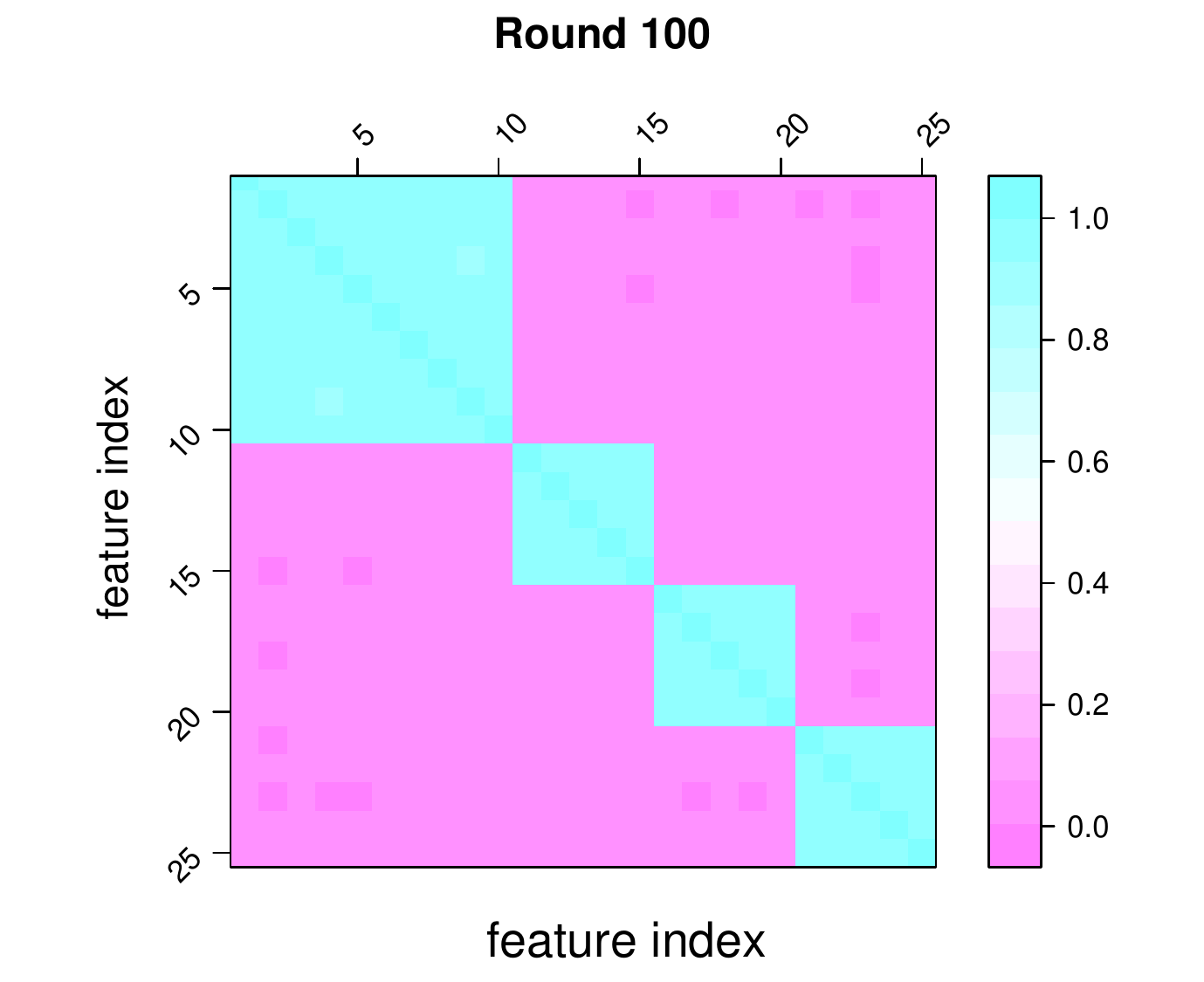}
		}\hspace{-0.2em}
		\caption{Evolution of $\bm{K}$ w.r.t.\ different number of feedbacks queried by the sequential model.}
		\label{Fig:3}
	\end{figure*}
	\subsection{Synthetic Data}
	%We conduct a simulation study to check if the proposed model behaves as expected and to compare the performance of different query models, i.e. random, non-sequential, and sequential.
	\emph{Data generation}: We simulate data from a linear regression model, $\bm{y} = \bm{X} \bm{\beta}+\bm{\epsilon}$, with $\bm{\epsilon} \sim \mathcal{N}(\bm{0},\sigma^2\bm{I})$ and $\sigma^2 = 5$. The data are generated similarly to \cite{li2010bayesian}, with the difference that instead of manually constructing the optimal regression coefficient, $\bm{\beta}_{opt}$, we construct the optimal $\bm{\gamma}$, denoted by $\bm{\gamma}_{opt}$, and the meta-features. They are used to construct the optimal covariance structure, $\bm{K}_{opt}$, using which we sample $\bm{\beta}_{opt}$ from $\mathcal{N}\left(\bm{0},\sigma^2\tau^2\bm{K}_{opt}\right)$, assuming $\tau^2 = 1$. The idea is to check whether or not the model can learn $\bm{\gamma}_{opt}$ and consequently $\bm{\beta}_{opt}$ using the feedback given by an oracle. We set the number of features to $25$ which are divided into four groups of sizes $10$, $5$, $5$, $5$. Features within each group are  highly correlated while features across groups are independent. Meta-features are vectors of size $9$ and dimensions $1$, $2$, $4$, $6$ are assumed to be important for learning the optimum covariance structure, i.e. $\bm{\gamma}_{opt} = \left[1,1,0,1,0,1,0,0,0\right]^T$. The optimum covariance matrix constructed using $\bm{\gamma}_{opt}$ and meta-features is shown in Figure \ref{Fig:3}.a..
	
	%By constructing $\bm{K}_{opt}$ and sampling $\bm{\beta}_{opt}$ from the prior distribution,
	Training and test outputs are generated by randomly sampling $\bm{X}_{tr}\in \mathbb{R}^{20\times 25}$ and $\bm{X}_{te}\in \mathbb{R}^{1000\times 25}$ from a standard normal distribution, multiplying them by $\bm{\beta}_{opt}$ and adding normally distributed noise with variance $\sigma^2$.
	
	\emph{Learning of the model and results}: The hyperparameters of the model are $a_{\sigma} = 2$ and $b_{\sigma} = 7$ for $\sigma^2$ and $a_{\tau} = 2$ and $b_{\tau} = 4$ for $\tau^2$, to reflect relatively vague information on the residual variance and scaling parameter, respectively. The prior for $\bm{\gamma}$ is set to $\mathcal{N}^{+}\left(1,0.5\right)$ meaning that \textit{a priori} we expect almost all dimensions of the meta-features to be equally important, i.e.,  $\bm{\gamma}_{0} \cong \left[1,1,1,1,1,1,1,1,1\right]^{T}$. In other words, any difference in any dimensions of the meta-feature will affect the (dis)similarity of the features. The hyperparameters of the threshold variable are $\mu_{\xi} = 20$ and $\sigma^{2}_{\xi} = 10$ to cover wide ranges of values. 
	
	The model was first trained to obtain an estimate of the regression coefficients without any feedback ($\bm{\beta}_0$). The posterior of the regression coefficients is then used to estimate the utility of each pair (to be used in the next round) and to compute the output of the test data. In each round, the simulated user gives one feedback for a pair queried using three different algorithms, i.e., random, non-sequential (using the utilities computed in round $0$) and sequential (updates utilities of each pair, except those to which the user already gave feedback, using estimated parameters in the previous round). Figures \ref{Fig:3}.b-d show how $\bm{K}$ evolves using feedback on pairs queried by the sequential model. The total number of possible feedbacks is $\frac{25\times 24}{2}=300$; using only one third of the feedbacks, the sequential model learned almost the optimal covariance structure.% The final covariance matrix (given all feedbacks; not shown here) is very similar to the one obtained with $100$ feedbacks which means that the rest of the pairs do not provide much extra information about groupings of the features.
	
\begin{figure}[b!]
		\centering
		% \subfloat[]{\includegraphics[scale=0.3]{beta_opt.pdf}} 
		% \subfloat[]{\includegraphics[scale=0.3]{beta_300.pdf}}\\
		% \subfloat[]{\includegraphics[scale=0.3]{mse_vs_100_feedback.pdf}} 
		% \subfloat[]{\includegraphics[scale=0.3]{mse_vs_all_feedback.pdf}}\\
		% \caption{(a, b) Comparison of $\bm{\beta}_{opt}$ and $\bm{\beta}_{100}$ obtained by sequential model and (c, d) predictive performances of different query algorithms} 
		\includegraphics[width = 0.6\columnwidth]{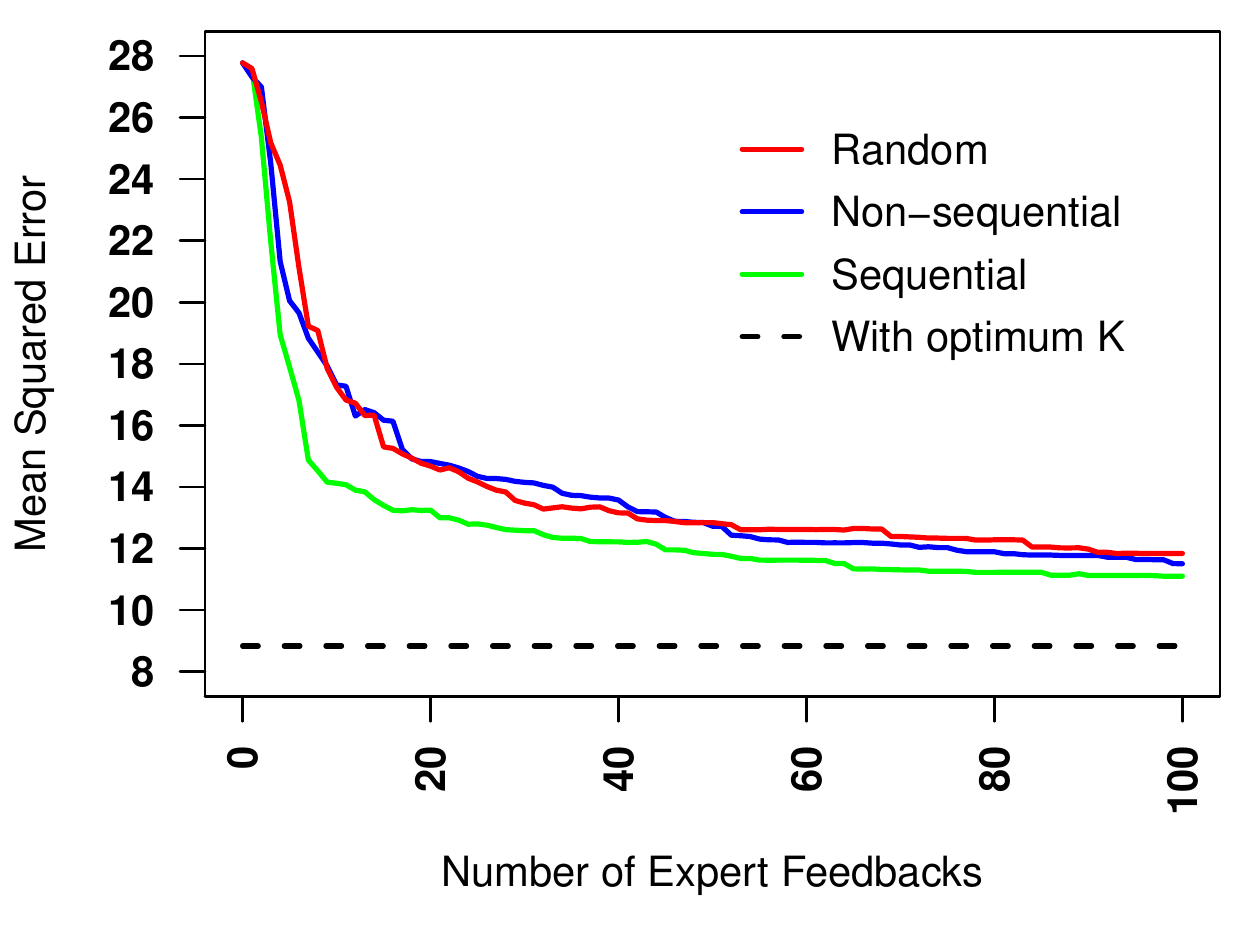}
		\caption{Comparison of predictive performances of different query algorithms in synthetic data.} 
		\label{Fig:4}
	\end{figure}

	We ran the model $50$ times with different randomly generated data with the setting explained above. The average predictive performances of the model with the different query algorithms are shown in Figure \ref{Fig:4}. The sequential model obtains the same amount of improvement faster than the random and non-sequential models. Both the random and non-sequential models also perform well, which is due to the simple structure of the data.
	\subsection{Real Data with Simulated Expert}
	
	We tested the proposed method in the task of review ratings prediction from textual reviews, using subsets of Amazon review data and Yelp data. Both data sets contain textual reviews represented as bag-of-words and their corresponding ratings, integer values in the range $ 1 - 5 $. Features correspond to keywords (unigrams and bigrams). For Amazon, we use the \textit{kitchen appliances} subset which contains $5149$ reviews \cite{blitzer2007biographies}. For \textit{Yelp}, we consider $4086$ reviews from the year 2004. In our analysis of both data sets, we removed stop words \cite{Salton1971SMART} and kept features that appeared in at least $100$ reviews. This resulted in data sets containing $340$ features for Amazon and $241$ features for Yelp. From each data set, $1000$ randomly selected reviews were used for training and tesing of the model and the rest as an ``expert data set'' for constructing the simulated expert. Among the $1000$ reviews for the model, $10\%$ were randomly selected for training and the remaining $90\%$ for testing. This resulted in a training set of size $100\times 340$ for Amazon and of size $100\times 241$ for Yelp, both of which small $n$, large $p$. Training data were normalized to have zero mean and unit variance. As meta-features, we used the transpose of the unnormalized training data. This is the simplest possible meta-feature according to which features (keywords) that are not in the same documents are dissimilar and vice versa. Better meta-features can be defined using word2vec \cite{mikolov2013distributed} or more advanced representation of the keywords; here we show that the model works fairly well even with such naive meta-features.
	
	\subsubsection{Simulated Expert Feedback}
	
	To construct the simulated expert, we trained a sparse regression model using LASSO \cite{friedman2010regularization} on the expert data set. Features with positive regression coefficients are considered pairwise similar. Those with negative coefficients are also similar, while features with zero weight are assumed to be difficult for the simulated expert to evaluate and the feedback will be ``I don't know'' for any pair containing at least one of those features. %Three groups of features are obtained by LASSO, i.e. features with (I) positive regression coefficients, (II) negative regression coefficients, and (III) zero regression coefficients. Features in groups $1$ and $2$ are considered pairwise similar and group-wise dissimilar, while features in group $3$ are difficult for the simulated expert to evaluate and the feedback will be ``I don't know'' for any pair containing at least one of those features.
	Since our goal is to evaluate the predictive performance of the models and feedback on these features has no effect on models' performance, we remove all pairs containing any of these features to make the runs faster. For both data sets, there are almost $20$ features in each of the two groups, resulting in $(40\times 39)/2 = 780$ pairs in total. Using proposition 1 for the sequential model, in each iteration  we randomly selected $400$ pairs among which the most informative one will be selected to query to the expert.
	
	The hyperparameters of the model were set to the same values as in the synthetic data, with the only difference that for both data sets we set $\tau^2 = 0.01$, obtained by cross-validation.
	
	\subsubsection{Results}
	
\begin{figure}[!]
		\centering
		\subfloat[Amazon]{\includegraphics[scale=0.4]{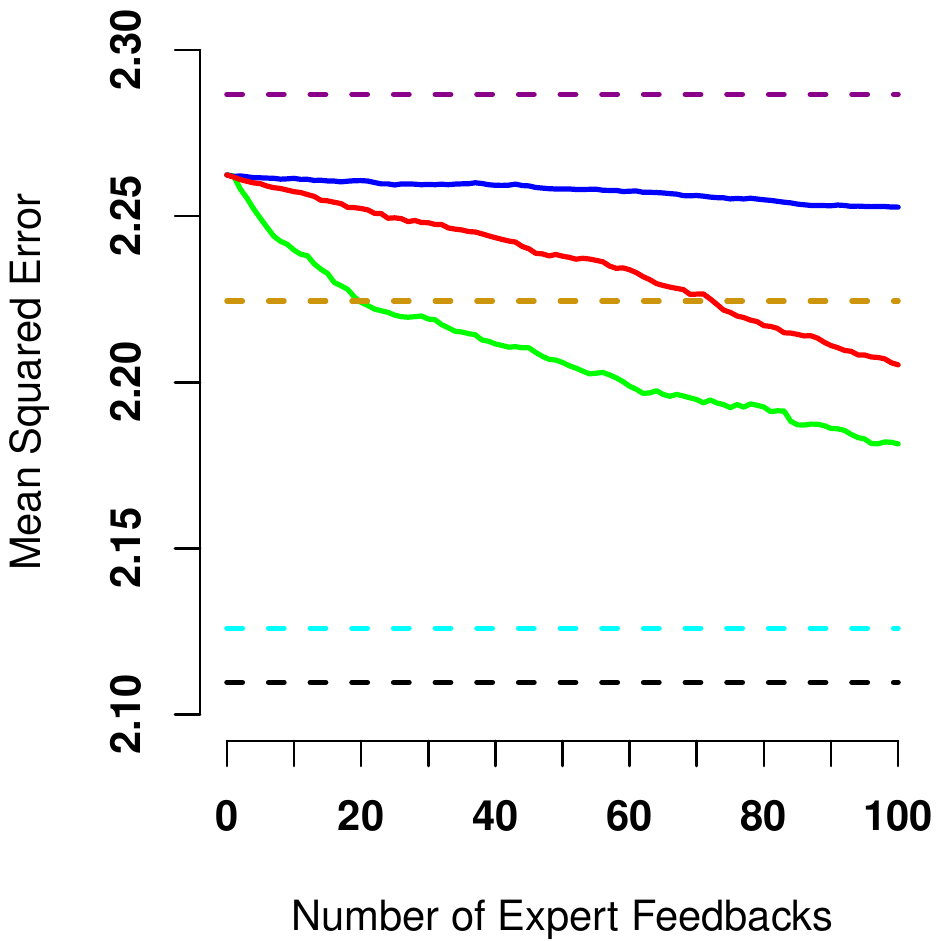}} 
		\subfloat[Yelp]{\includegraphics[scale=0.4]{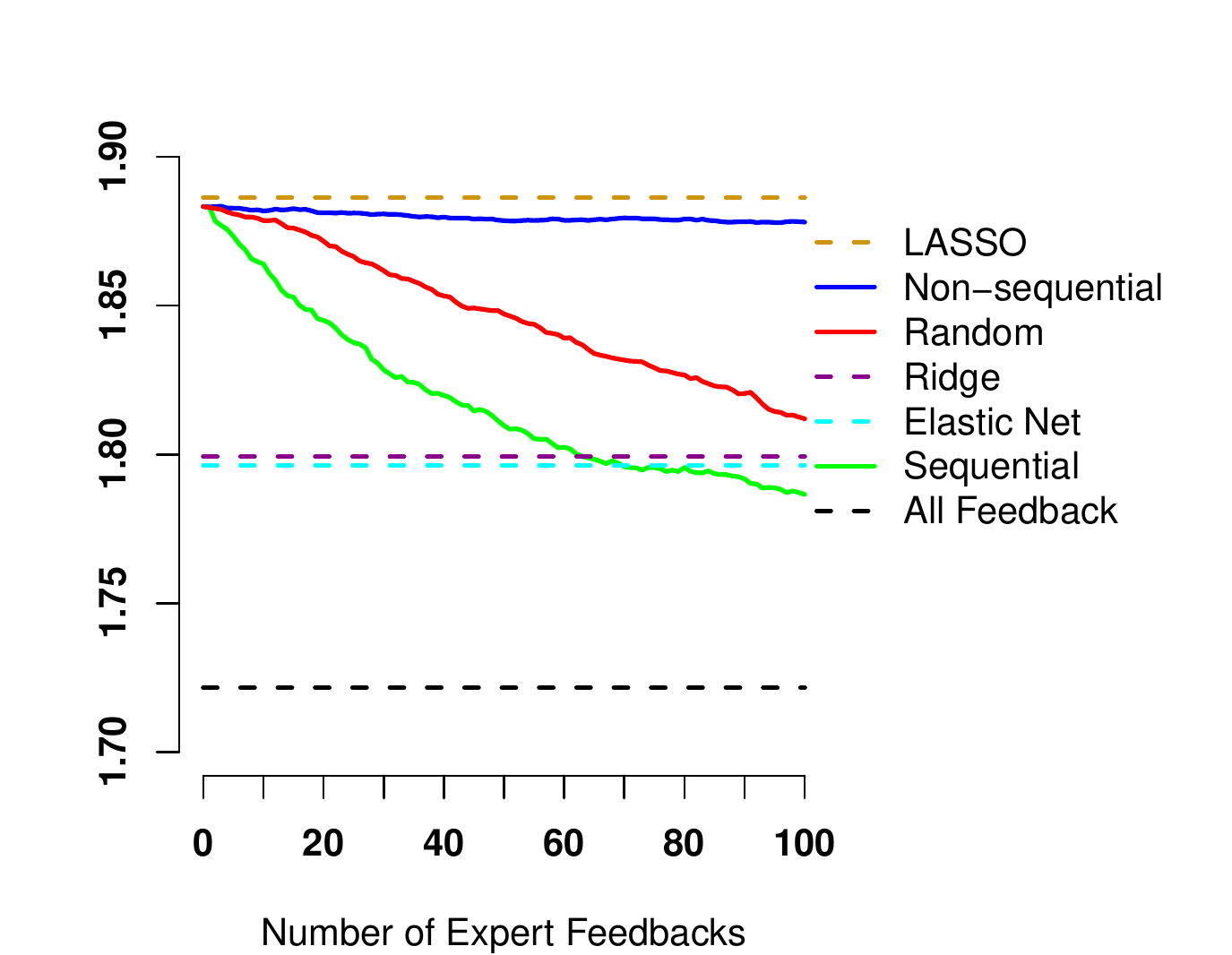}}
		\caption{Mean squared error w.r.t.\ the number of feedbacks on pairs of features. MSE values are averages over 50 independent runs.} 
		\label{Fig:5}
	\end{figure}

	Figure \ref{Fig:5} shows the predictive performance of query algorithms on each data set. Results are averaged over $50$ runs with randomly selected training, test, and expert data. Query algorithms ask one question at a time and the model was ran for $100$ rounds. % The ground truth line represents the MSE after receiving all feedbacks and baseline is the performance of the model without feedback (iteration $0$ of our method).
	For all algorithms and on both data sets, additional knowledge from the simulated expert reduces the prediction error. Yet the amount of reduction depends on the query algorithm. The proposed method in sequential mode obtains fastest improvement. The random model performs better than the non-sequential model. This is expected since the non-sequential model is prone to querying redundant pairs; for instance when there is a cluster of similar features that are highly informative, %in constructing the prior covariance,
	asking a few of them is sufficient to inform of their grouping. However, the non-sequential model will query all the possible pairs since it does not update the utilities. To compare the effect of learned covariances using expert feedback on regularizing the model, we also compare to the baselines that make a sparsity assumption to regularize the model: LASSO, Ridge and Elastic net. It can be seen that a relatively small number of expert feedback could regularize the model well and achieve good prediction performance compared to the commonly used regularization methods.
	\subsubsection{Knowledge Elicitation vs. Collecting More Samples}
	Table \ref{T1} compares the number of pairwise feedback required to obtain the same MSE value achieved by adding additional samples (reviews from the expert data set) to the training set. For the latter, we use two strategies: random selection of samples (shown by \textbf{\textbf{RND}}), and active learning strategy (shown by \textbf{\textbf{ACT}}), which selects samples based on maximizing expected information gain (similar to \cite{seeger2008bayesian}). For feedback collection, we adopt our sequential knowledge elicitation approach (shown by \textbf{\textsc{SeqElc}}). All strategies have the same ``small $n$, large $p$'' setting as starting points with $\bm{X}_{tr} \in \mathbb{R}^{100\times 241}$ for Yelp and $\bm{X}_{tr} \in \mathbb{R}^{100\times 340}$ for Amazon.
	
	The comparison shows the potential of expert knowledge on feature similarities in the prediction when obtaining more samples is impossible or very expensive. According to the table, a particular performance is obtained by a comparable number of expert feedback and additional data. It should be noted that values obtained for the sequential elicitation model are for its best case scenario since we already removed ``I don't know''s which results in a much smaller search pool.
	\begin{table}[ht!]
		\centering
		\caption{Comparison of the number of required samples and feedbacks to reach a particular MSE level. Values are averages over 50 independent runs. Initial MSE values for Yelp and Amazon are $1.88$ and $2.26$, respectively.}\label{T1}
		\begin{tabular}{ccccc}
			& & \multicolumn{2}{|c|}{More Samples} & More Feedbacks \\ \cline{2-5} 
			\multicolumn{1}{c}{} & \multicolumn{1}{c|}{MSE} & \multicolumn{1}{c|}{\textbf{\textbf{RND}}} & \multicolumn{1}{c|}{\textbf{\textbf{ACT}}} & \multicolumn{1}{c}{\textbf{\textsc{SeqElc}}} \\ \hline
			\multicolumn{1}{c||}{\multirow{2}{*}{Yelp}} & \multicolumn{1}{c|}{1.85} & \multicolumn{1}{c|}{5} & \multicolumn{1}{c|}{5} & \multicolumn{1}{c}{13} \\  \multicolumn{1}{c||}{} & \multicolumn{1}{c|}{1.80} & \multicolumn{1}{c|}{15} & \multicolumn{1}{c|}{9} & \multicolumn{1}{c}{60} \\ \hline
			\multicolumn{1}{c||}{\multirow{2}{*}{Amazon}} & \multicolumn{1}{c|}{2.20} & \multicolumn{1}{c|}{20} & \multicolumn{1}{c|}{14} & \multicolumn{1}{c}{55} \\  
			\multicolumn{1}{c||}{} & \multicolumn{1}{c|}{2.18} & \multicolumn{1}{c|}{27} & \multicolumn{1}{c|}{17} & \multicolumn{1}{c}{90} \\ \hline
		\end{tabular}
	\end{table}
	\section{Discussion and Conclusion}
	
	We proposed a knowledge elicitation approach that incorporates pairwise constraints provided by a human expert for constructing an informative prior distribution. %The constructed prior distribution allows us to learn the dependence structure from feedback on the (dis)similarities of pairs of features.
	The knowledge elicitation problem is formulated as sequential probabilistic inference which combines expert knowledge with training data. To avoid overwhelming the expert, we proposed a query selection approach that facilitates efficient interaction. To further increase the efficiency of the interaction, we formulated a theoretically motivated subsampling approach. Results for ``small $n$, large $p$'' problems in simulated and real data with simulated users showed improved prediction accuracy already with a small number of feedbacks.
	
	From our experience, the proposed algorithm works well for moderately high-dimensional problems (e.g., up to 500 predictors). For very high-dimensional problems, with more than thousand predictors, the computations become infeasible due to the large number of pairs. However, the presented approach is general and can be used to complement other existing knowledge elicitation techniques where the user is asked about relevance of features. In this case we can query only pairs of relevant features which significantly improves the feasibility of the model.% It could also be used together with a suitable visualization method such as those in \cite{afrabandpey2017interactive,he2017medisyn}. A potential future direction is to investigate extension of this work to different types of interactive machine learning problems.
	\section*{Acknowledgments}
	This work was financially supported by the Academy of Finland (grants 294238, 319264, and 313195), by the  Vilho, Yrj\"{o} and Kalle V\"{a}is\"{a}l\"{a} Foundation of the Finnish Academy of Science and Letters and by Foundation for Aalto University Science and Technology. The authors acknowledged the computational resources provided by the Aalto Science-IT Project.
	
	\bibliographystyle{named}
	\bibliography{References.bib}
	
	\newpage
	\onecolumn
	\section*{Appendix A: KL divergence of predictive distributions}
	In the following formulas, distributions are shown by $p(.)$ in contrast to the main text where we used $\pi(.)$. Let us assume that the predictive distribution before receiving a new feedback $f$ is shown by
	\begin{flalign*}
		p(\Tilde{y} \mid y) = t_{2a^{*}}\left( \Tilde{\bm{x}}^{T}\bm{\mu}^{*}, \frac{b^{*}}{a^{*}}\left( 1 + \Tilde{\bm{x}}^{T}\bm{\Sigma}^{*}\Tilde{\bm{x}} \right) \right)
	\end{flalign*}
	where $\Tilde{\bm{x}}$ is the future input with its corresponding output $\Tilde{y}$ and $a^*$, $b^*$, $\bm{\mu}^{*}$ and $\bm{\Sigma}^*$ are defined in Section 3.5. The predictive distribution after observing a new feedback $f$ is shown by
	\begin{flalign*}
		p(\Tilde{y} \mid y, f) = t_{2a^{*}}\left( \Tilde{\bm{x}}^{T}\bm{\mu}_{new}^{*}, \frac{b_{new}^{*}}{a^{*}}\left( 1 + \Tilde{\bm{x}}^{T}\bm{\Sigma}_{new}^{*}\Tilde{\bm{x}} \right) \right).
	\end{flalign*}
	Note that, each feedback updates the point estimates of $\tau^2$ and $\bm{K}$ and only those variables depend on these two parameters will be updated, i.e. $b^{*}$, $\bm{\mu}^{*}$, and $\bm{\Sigma}^{*}$. The updated values are shown by $b_{new}^{*}$, $\bm{\mu}_{new}^{*}$, and $\bm{\Sigma}_{new}^{*}$, respectively (Section \ref{Sec:4.3}).
	
	For the sake of computational simplicity, we relax the above student's t-distributions to univariate Gaussian distributions. This is possible particularly due the large degree of freedoms of the two distributions (note that $a^{*} = a_{\sigma} + \frac{N}{2}$ where $N$ is the number of samples). Approximations are as follows:
	\begin{flalign}
		\begin{split}
			p(\Tilde{y} \mid y) = \mathcal{N}\left( \Tilde{\bm{x}}^{T}\bm{\mu}^{*}, \frac{b^{*}}{a^{*}-1}\left( 1 + \Tilde{\bm{x}}^{T}\bm{\Sigma}^{*}\Tilde{\bm{x}} \right) \right)
		\end{split}\\
		\begin{split}
			p(\Tilde{y} \mid y, f) = \mathcal{N}\left( \Tilde{\bm{x}}^{T}\bm{\mu}_{new}^{*}, \frac{b_{new}^{*}}{a^{*}-1}\left( 1 + \Tilde{\bm{x}}^{T}\bm{\Sigma}_{new}^{*}\Tilde{\bm{x}} \right) \right).
		\end{split}
	\end{flalign}
	With these approximations, the KL-divergence of the two Gaussian distributions are obtained as
	\begin{multline}
		\mbox{KL}\left( p(\Tilde{y} \mid y, f) \parallel p(\Tilde{y} \mid y) \right) = \log \left( \sqrt{\frac{b^{*}\left( 1 + \Tilde{\bm{x}}^{T}\bm{\Sigma}^{*}\Tilde{\bm{x}} \right)}{b_{new}^{*} \left( 1 + \Tilde{\bm{x}}^{T}\bm{\Sigma}_{new}^{*}\Tilde{\bm{x}} \right)}} \right) + \frac{(a^{*}-1)\left( \Tilde{\bm{x}}^{T}\left( \bm{\mu}_{new}^{*} - \bm{\mu}^{*} \right) \right)^{2}}{2b^{*}\left( 1 + \Tilde{\bm{x}}^{T}\bm{\Sigma}^{*}\Tilde{\bm{x}} \right)}\\
		+ \frac{b_{new}^{*} \left( 1 + \Tilde{\bm{x}}^{T}\bm{\Sigma}_{new}^{*}\Tilde{\bm{x}} \right)}{2b^{*}\left( 1 + \Tilde{\bm{x}}^{T}\bm{\Sigma}^{*}\Tilde{\bm{x}} \right)} - \frac{1}{2}
	\end{multline}
	Note that since we are using the training data to compute the utilities of the queries, in the implementation, each training data is used as the unobserved input $\Tilde{\bm{x}}$ and the KL-divergence is computed as the summation over all the training data.
	\newpage
	\onecolumn
	\section*{Appendix B: Marginal likelihood derivation}
	\begin{flalign*}
		\begin{split}
			p(\bm{y}) = \int p\left( \bm{y} \mid \bm{\beta}, \sigma^{2} \right)p\left( \bm{\beta} \mid \sigma^{2} \right)p(\sigma^{2}) d\bm{\beta}d\sigma^{2} = \int \mathcal{N}\left( \bm{y} \mid \bm{X\beta}, \sigma^{2}\bm{I} \right) \mathcal{N}\left( \bm{\beta} \mid \bm{0}, \sigma^{2}\tau^{2}\bm{K} \right) \mbox{IG}\left( \sigma^{2} \mid a_{\sigma},b_{\sigma}  \right) d\bm{\beta}d\sigma^{2}
		\end{split} \\
		\begin{split}
			= \int \left(2\pi \sigma^{2} \right)^{-\frac{N+M}{2}} \mid  \tau^{2}\bm{K} \mid^{-1/2} \mbox{exp} \left[-\frac{1}{2\sigma^{2}}\{ \left( \bm{y}-\bm{X\beta} \right)^{T}\left( \bm{y}-\bm{X\beta} \right) + \bm{\beta}^{T}\left( \tau^{2}\bm{K} \right)^{-1}\bm{\beta} \} \right]d\bm{\beta}\mbox{IG}\left(\sigma^{2} \mid a_{\sigma},b_{\sigma} \right)d\sigma^{2}
		\end{split} \\
		\begin{split}
			= \int \left(2\pi \sigma^{2} \right)^{-\frac{N}{2}} \mid  \bm{I}+\bm{X}\tau^{2}\bm{KX}^{T} \mid^{-1/2} \mbox{exp} \left[-\frac{1}{2\sigma^{2}}\{ \bm{y}^{T}\left( \bm{I}+\bm{X}\tau^{2}\bm{K}\bm{X}^{T} \right)^{-1}\bm{y} \} \right] \mbox{IG}\left( \sigma^{2} \mid a_{\sigma},b_{\sigma} \right)d\sigma^{2}
		\end{split} \\
		\begin{split}
			= \int \mathcal{N}\left( \bm{y} \mid \bm{0}, \sigma^{2}\left( \bm{I} + \tau^{2}\bm{XKX^{T}} \right) \right)\mbox{IG}\left( \sigma^{2} \mid a_{\sigma},b_{\sigma} \right)d\sigma^{2}
		\end{split}
	\end{flalign*}
	To make the following equations shorter, we define $\bm{\Sigma} = \bm{I} + \tau^{2}\bm{XK}\bm{X}^{T}$, then we can write:
	\begin{flalign*}
		\begin{split}
			\int (2\pi)^{-\frac{N}{2}}(\sigma^{2})^{-\frac{N}{2}} \mid \bm{\Sigma} \mid^{-1/2} \mbox{exp}\left( -\frac{1}{2\sigma^{2}}\bm{y}^{T} \bm{\Sigma}^{-1}\bm{y} \right) \frac{b_{\sigma}^{a_{\sigma}}}{\Gamma(a_{\sigma})}( \sigma^{2})^{- a_{\sigma} - 1} \mbox{exp}( -\frac{b_{\sigma}}{\sigma^{2}}) d\sigma^{2}
		\end{split} \\
		\begin{split}
			= (2\pi)^{-\frac{N}{2}} \frac{b_{\sigma}^{a_{\sigma}}}{\Gamma(a_{\sigma})} \mid \bm{\Sigma} \mid^{-1/2} \int (\sigma_{2})^{-\frac{N}{2} - a_{\sigma} - 1} \mbox{exp}\left( -\frac{1}{\sigma^{2}}\left[ b_{\sigma} + \frac{1}{2} \bm{y}^{T}\bm{\Sigma}^{-1} \bm{y}\right] \right)d\sigma^{2}
		\end{split} \\
		\begin{split}
			= (2\pi)^{-\frac{N}{2}}\frac{\Gamma \left( a_{\sigma}+ \frac{N}{2}\right)}{\Gamma(a_{\sigma})}b_{\sigma}^{a_{\sigma}}\mid \bm{\Sigma} \mid^{-1/2}\left( b_{\sigma} + \frac{1}{2} \bm{y}^{T}\bm{\Sigma}^{-1} \bm{y} \right)^{-\left( a_{\sigma} + \frac{N}{2} \right)}
		\end{split}
	\end{flalign*}
	where $\Gamma(.)$ and $\mid . \mid$ represent the Gamma function and the determinant, respectively. The last term is the form of a multivariate student's t-distribution with $2a_{\sigma}$ degrees of freedom and parameters $\bm{0}$ and $\frac{b_{\sigma}}{a_{\sigma}}\bm{\Sigma}$.
	\newpage
	\onecolumn
	\section*{Appendix C: Plate Diagram of the Linear Regression Model}
	Plate diagram of the prediction model (left) and feedback model (right). Feedbacks $f_l$ are sequentially queried from the expert where the total number of feedbacks is equal to the number of feature pairs.
	\begin{figure}[h!]
		\centering
		\includegraphics[width = 0.8\columnwidth]{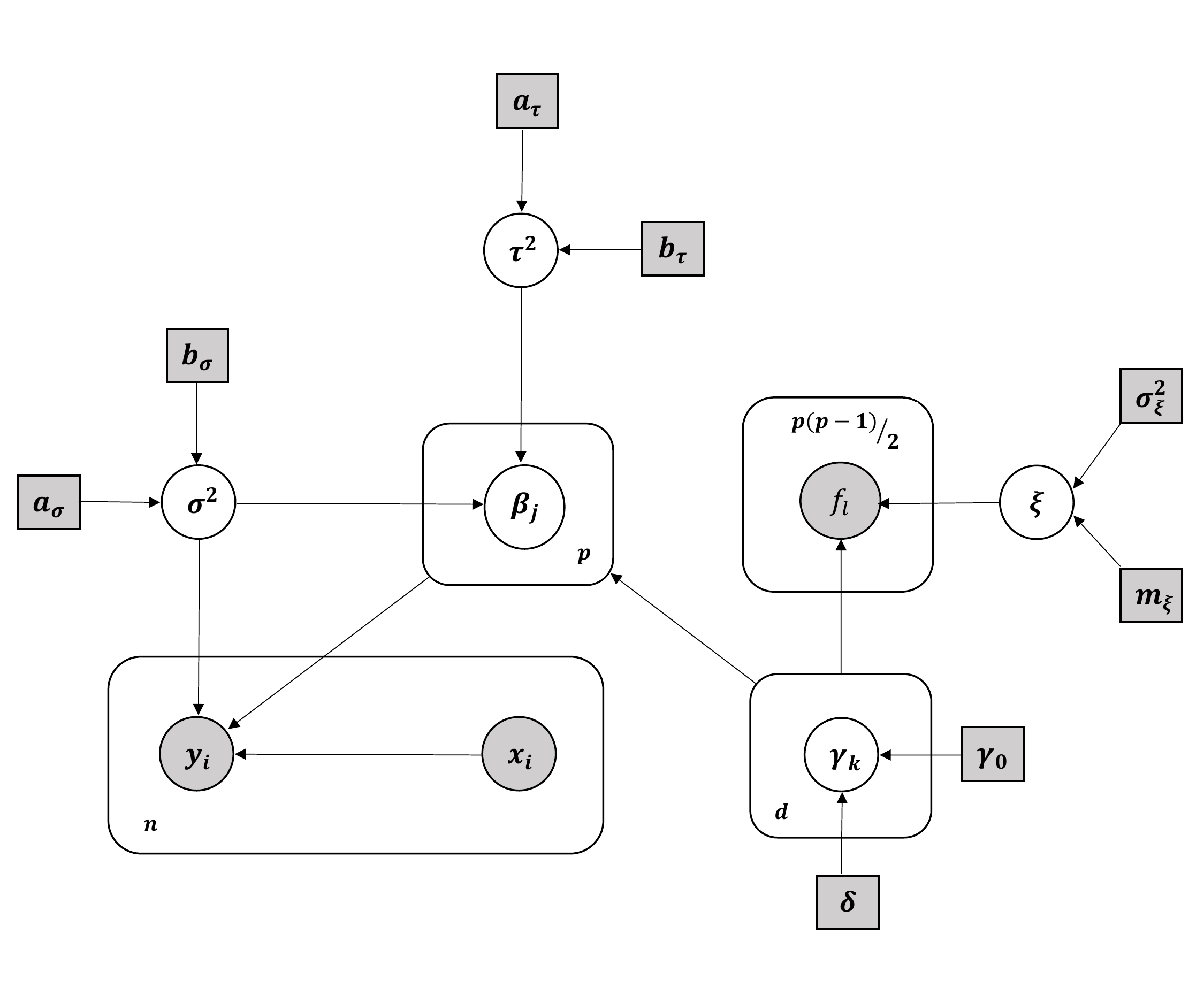}
		% \caption{Caption}
		\label{fig:my_label}
	\end{figure}

	\newpage
	\onecolumn
	\section*{Stan Codes}
	\lstset{language=C, tabsize=2, showstringspaces=false,
		basicstyle=\ttfamily\scriptsize, literate={~} {$\sim$}{1}} 
\begin{lstlisting}[language=C++]
data{
	int N;									// number of observations
	int M;									// number of predictors
	int N_f;								// number of pairs for which the user already gave feedback
	int d;									// length of the feature descriptor
	vector[N] y;						// outputs
	matrix[N, M] X;					// observed data matrix
	matrix[M, d] FD;				// matrix of feature descriptors
	real a_sigma;						// shape of the residual noise distribution
	real b_sigma;						// scale of the residual noise distribution
	
	// scaler value; in case of non-constant tau, you need 'real a_tau' and 'real b_tau'
	real tau;
	
	// feedback vector with values 1/0; 1 = similarity and 0 = dissimilarity
	int<lower = 0, upper = 1> f[N_f];
	
	// pairwise distance vectors of pairs to which the user gave feedback
	matrix[N_f, d] W_f;
	real threshold_mean;			// initial value of the threshold \mu
}
parameters{
	vector<lower=0>[d] gamma;	// diagonal elements of the metric matrix A
	real<lower=0> threshold;
}
transformed parameters{
	cov_matrix[M] K;					// covariance structure
	vector[N_f] q;						// feedback likelihood
	
	K = quad_form_sym(diag_matrix(gamma), FD');
	K = exp(-0.5 * (rep_matrix(diagonal(K), M) + rep_matrix(diagonal(K)', M)) + K);
	for(m in 1:M) K[m, m] = K[m, m] + 1e-06;
	q = 1.0 ./ (1.0 + exp(W_f * gamma - threshold));
}
model{
	\\	half-normal for mu
	threshold ~ normal(threshold_mean, threshold_mean/2);
	
	\\	half-normal for gamma
	gamma ~ normal(1.0, 0.5);
	
	\\	bernoulli for feedback likelihood
	f ~ bernoulli(q);
	
	{
		matrix[N, N] S;
		S = tau * quad_form_sym(K, X');
		for(n in 1:N) S[n, n] = S[n, n] + 1.0;
		
		\\ multivariate student t distribution for marginal likelihood
		y ~ multi_student_t(2.0 * a_sigma, rep_vector(0, N),(b_sigma / a_sigma) * S);
	}
}
\end{lstlisting}
	
\end{document}